\documentclass[sigconf,natbib=true]{acmart}
\AtBeginDocument{%
  }

\usepackage{hyperref}       % hyperlinks
\usepackage{amsfonts}       % blackboard math symbols
\usepackage{bm}
\usepackage{graphicx}
\usepackage{amsthm}
\usepackage{amsmath}
\usepackage{subfigure}
\usepackage{multirow}
\usepackage{enumitem}
\usepackage{bbding}
\usepackage{colortbl}
\usepackage{arydshln}
\usepackage{algorithm}  
\usepackage{algorithmicx}  
\usepackage{threeparttable}
\usepackage{algpseudocode}

\setcopyright{acmlicensed}
\copyrightyear{2018}
\acmYear{2018}
\acmDOI{XXXXXXX.XXXXXXX}
\acmConference[Conference acronym 'XX]{Make sure to enter the correct
  conference title from your rights confirmation emai}{June 03--05,
  2018}{Woodstock, NY}
\acmISBN{978-1-4503-XXXX-X/18/06}

\begin{document}

% \title{ReSLR: Enabling Explicit Syllogistic Legal Reasoning in LLM \\via Reinforcement Fine-Tuning}
\title{
An Explicit Syllogistic Legal Reasoning Framework \mbox{for Large Language Models}
}

\author{Kepu Zhang}
\affiliation{
  \institution{\mbox{Gaoling School of Artificial Intelligence}\\Renmin University of China} \city{Beijing}\country{China}
  } 
\email{kepuzhang@ruc.edu.cn}

\author{Weijie Yu}
\affiliation{
  \institution{School of Information Technology 
and Management\\University of International Business
and Economics} \city{Beijing}\country{China}
  }

\author{Zhongxiang Sun}
\author{Jun Xu}
\affiliation{
  \institution{\mbox{Gaoling School of Artificial Intelligence}\\Renmin University of China} \city{Beijing}\country{China}
  }

% \affiliation{%

% \affiliation{%

\renewcommand{\shortauthors}{Kepu Zhang et al.}

\begin{abstract}
Syllogistic reasoning is crucial for sound legal decision-making, allowing legal professionals to draw logical conclusions by applying general principles to specific case facts. While large language models (LLMs) can answer legal questions, they often struggle with explicit syllogistic reasoning. Their outputs tend to be implicit, unstructured, and consequently, less explainable and trustworthy.
To overcome these limitations, we introduce SyLeR, a novel framework designed to enable LLMs to perform explicit syllogistic legal reasoning. SyLeR employs a tree-structured hierarchical retrieval mechanism to synthesize relevant legal statutes and precedents, thereby constructing comprehensive major premises. This is followed by a two-stage fine-tuning process: an initial supervised fine-tuning warm-up establishes a foundational understanding of syllogistic reasoning, while reinforcement learning, guided by a structure-aware reward mechanism, refines the model's capacity to generate diverse, logically sound, and well-structured reasoning paths.
We conducted extensive experiments to evaluate SyLeR's performance. Our evaluations spanned diverse dimensions, including both in-domain and cross-domain user groups (legal laypersons and practitioners), multiple languages (Chinese and French), and various LLM backbones (legal-specific and open-domain LLMs). The results consistently demonstrate that SyLeR significantly enhances response accuracy and reliably produces explicit, explainable, and trustworthy legal reasoning.
\end{abstract}





\maketitle

\section{Introduction}\label{sec:intro}
\begin{figure}
    \centering
\includegraphics[width=0.98\linewidth]{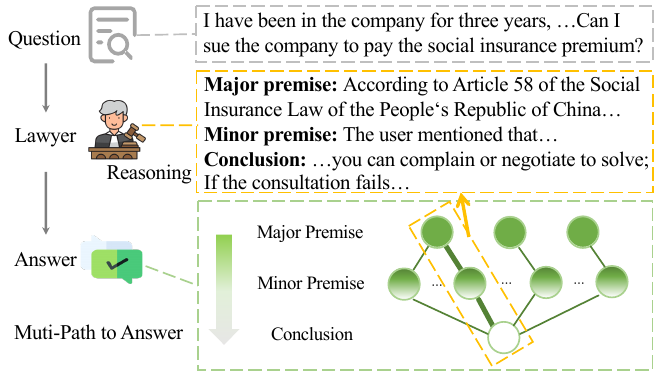}
    \caption{
    Syllogistic reasoning in legal QA: The lawyer identifies the relevant legal statutes or precedent cases to derive the major premise, formulates the minor premise from the user's question, and then reaches the conclusion. Based on different analysis perspectives, different syllogistic reasoning paths can be derived.
            }
    \label{fig:fig 1}
    % \vspace{-3mm}
\end{figure}

Legal decision-making fundamentally depends on logical reasoning to ensure fairness and trustworthy in judgments. Among the various forms of legal reasoning, syllogistic reasoning~\cite{deng2023syllogistic,jiang2023legal} stands out as a structured form of deductive reasoning that derives conclusions from two interconnected premises. In legal contexts, the major premise typically comprises general legal rules or principles—such as statutes and precedents—while the minor premise is drawn from the specific facts of a case. The logical connection between these premises leads to a well-founded legal conclusion. This reasoning process closely reflects how legal professionals analyze cases, ensuring that decisions are both grounded in established laws and logically aligned with case facts, Figure~\ref{fig:fig 1} is an example.

Despite the fundamental role of syllogistic reasoning in legal decision-making, existing large language models (LLMs) struggle to replicate this explicit reasoning process in legal tasks. On the one hand, legal-specific LLMs~\cite{yue2023disclawllm,sdu_fuzi_mingcha,wisdomInterrogatory} have primarily focused on enhancing legal knowledge through supervised fine-tuning on domain-specific datasets. These models not only require a substantial amount of annotated data but also lack clear, logically structured explanations that mirror how legal professionals connect statutes and case facts to reach decisions. This limitation undermines both the explainability and trustworthiness of legal LLMs, hindering their practical deployment in scenarios where transparent and well-justified reasoning is critical. 
On the other hand, research aimed at enhancing the reasoning capabilities of LLMs has introduced methods such as Chain-of-Thought (CoT) prompting~\cite{shum2023automatic,wei2022chain}, 
% In-Context Learning (ICL)~\cite{wei2022chain}, 
supervised fine-tuning~\cite{herevisiting}, and retrieval-augmented fine-tuning~\cite{zhang2024raft}. 
These techniques enhance the model's reasoning ability by either implicitly guiding the model or explicitly adjusting the model parameters, thereby
% These techniques guide models to decompose complex problems into intermediate reasoning steps, 
improving performance in tasks involving mathematics, commonsense reasoning, and code generation. However, these approaches remain largely domain-agnostic and fail to address the distinct challenges of legal reasoning, particularly the need for generating explicit syllogistic reasoning paths that align legal rules with case-specific facts.

Enabling LLMs to perform explicit syllogistic legal reasoning involves addressing several complex challenges.
\textbf{First}, effectively incorporating comprehensive legal knowledge into LLMs is inherently difficult. Legal reasoning depends on both statutes and precedent cases~\cite{zhang2024citalaw}, which are not only diverse in form but also deeply interconnected. Statutes provide general legal principles, while precedents interpret and apply these statutes to specific cases. 
Treating these sources separately can disrupt their contextual relationships, making it difficult for models to capture the nuanced interplay between legal rules and their applications. Therefore, designing models that can holistically incorporate and reason over both statutes and precedents within a structured framework is essential.
\textbf{Second}, achieving explicit syllogistic reasoning requires the model to learn how to construct valid reasoning paths. However, for any given legal question, there may exist multiple valid syllogistic reasoning paths leading to the same conclusion, as shown in Figure~\ref{fig:fig 1}. Annotating diverse reasoning paths for supervised fine-tuning is expensive and labor-intensive, particularly in the legal domain where expert annotation is required. More importantly, how to train LLMs to learn from and generalize across multiple reasoning strategies—without overfitting to a single annotated path—is a complex challenge that must be addressed.
\textbf{Third}, evaluating an LLM’s performance in syllogistic legal reasoning demands more than standard text generation metrics. Conventional metrics such as accuracy and ROUGE fall short in assessing the logical consistency, coherence, and legal validity of the model’s reasoning processes. Without specialized evaluation criteria, it is difficult to determine whether a model truly understands and applies syllogistic reasoning. Therefore, it is crucial to develop task-specific evaluation methods that can effectively measure the soundness and transparency of the model’s reasoning chains, ultimately guiding the model toward producing more trustworthy and explainable legal outcomes. 

To address these challenges, we propose \textbf{SyLeR}, a novel framework designed to enable LLMs to perform explicit \textbf{Sy}llogistic \textbf{Le}gal \textbf{R}easoning. SyLeR integrates a two-stage training process that systematically incorporates legal knowledge and guides the model to generate logically structured reasoning. Specifically, we first introduce a tree-structured hierarchical retrieval mechanism that organizes legal knowledge by connecting relevant statutes and their corresponding precedent cases. This structure preserves the inherent relationship between legal rules and their practical applications, allowing the model to construct a comprehensive major premise that reflects both legal principles and real-world interpretations.
Building on this, SyLeR employs a two-stage reinforcement fine-tuning strategy. In the warm-up stage, we utilize GPT-4o to generate high-quality syllogistic reasoning paths by reverse engineering from retrieved legal documents, user questions, and ground-truth answers. These reasoning paths explicitly follow the structure of major premise, minor premise, and conclusion, providing the foundation for supervised fine-tuning that teaches the model how to produce logically coherent and structured responses.
In the reinforcement learning (RL) stage, we refine the model’s reasoning ability using the Proximal Policy Optimization (PPO)~\cite{schulman2017proximal} algorithm. This stage allows the model to explore diverse syllogistic reasoning paths, addressing the challenge of reasoning diversity. A carefully designed reward function evaluates the model’s output based on the alignment between the generated major premise and the retrieved legal knowledge, the consistency of the minor premise with the user question, and the correctness of the conclusion. To ensure strict adherence to the syllogistic reasoning format, responses that deviate from the expected structure receive zero rewards, guiding the model toward generating explicit and legally sound reasoning.

We conducted extensive experiments to evaluate SyLeR across multiple dimensions, including datasets from legal laypersons and practitioners, multiple languages (Chinese and French), and comparisons between legal-specific and open-domain LLMs. Additionally, cross-domain tests were performed by training on one user group and evaluating on another to assess adaptability. The results show that SyLeR significantly improves answer accuracy and provides explicit syllogistic reasoning, ensuring both trustworthiness and explainability across diverse legal scenarios.

In summary, the contributions of this paper are as follows:  
\begin{itemize}
    \item We pioneered the explicit incorporation of syllogistic legal reasoning into LLMs through reinforcement fine-tuning, enabling the LLM to generate responses in the format of major premise, minor premise, and conclusion.  
    \item We propose SyLeR, a novel framework that achieves explicit syllogistic legal reasoning through a tree-structured hierarchical retrieval mechanism for legal knowledge integration and a two-stage fine-tuning process. This process combines supervised learning for foundational reasoning and reinforcement learning with a structure-aware reward mechanism that evaluates the alignment of the major premise, minor premise, and conclusion, ensuring logically sound and structured outputs.  
    \item We conduct extensive experiments across diverse settings including different user groups (legal laypersons and practitioners), multiple languages (Chinese and French), and model types (legal-specific and open-domain LLMs) demonstrating that SyLeR significantly improves response accuracy and provides explicit and trustworthy legal reasoning across various legal contexts.
\end{itemize}

\section{Related Work}\label{sec:re}
\subsection{Legal LLM}
With the rise of LLMs, a large number of studies investigating the use of LLMs in the legal domain have emerged~\cite{jiang2023legal,choi2021chatgpt}. Early works explored the capabilities of LLMs in various legal tasks, using prompt-based methods~\cite{yu2022legal,trautmann2022legal} to provide examples to guide the LLM in solving tasks, including legal judgment prediction, legal examinations, and more. Some works have also designed benchmarks~\cite{fei2023lawbench,lilexeval} to test the capabilities of LLMs in legal tasks, evaluating them from multiple dimensions.
As publicly available legal-related data resources have become more abundant, constructing fine-tuning and training data based on these datasets has become easier. As a result, many studies have fine-tuned open-domain LLMs on large amounts of legal-related data, injecting legal knowledge into the models to create LLMs specifically for the legal field, enabling them to tackle legal-related tasks. For instance, \cite{yue2023disclawllm} fine-tuned LLM using high-quality supervised datasets constructed from judgment prediction, information extraction, and legal QA datasets. \cite{wisdomInterrogatory} conducted secondary pre-training and instruction fine-tuning on open-domain LLMs using datasets such as legal documents and legal cases. \cite{LAWGPT-zh} obtained high-quality judicial QA data through ChatGPT for fine-tuning. \cite{HanFei} performed full-parameter fine-tuning to create legal-specific LLMs.
Despite the promising results achieved by these legal LLMs on legal tasks, the final answers they provide are still a form of implicit reasoning. None of these methods have enabled the LLMs to perform explicit legal reasoning, which affects the trustworthiness of the LLMs and hinders their deployment in real-world scenarios.

\subsection{Reasoning in LLM}
Existing work has explored various strategies to enhance the reasoning capabilities of large language models (LLMs). One such strategy is the Chain-of-Thought (CoT) prompting technique~\cite{zhangautomatic,shum2023automatic}, which encourages the model to generate a series of intermediate reasoning steps through specific prompts. For example, the simple prompt "Let's think step by step"~\cite{kojima2022large} has been shown to improve the reasoning ability of LLMs. Further, In-Context Learning (ICL)~\cite{wei2022chain} involves providing examples containing intermediate reasoning steps to teach the model how to perform reasoning.
In the retrieval-augmented generation scenarios, enhancing the LLM's reasoning ability can be achieved by improving the prompts input to the LLM using its feedback, including compressing the retrieved documents~\cite{li2023compressing,jiang2023longllmlingua} and iteratively retrieval to augment generation~\cite{shao2023enhancing}.
Additionally, supervised fine-tuning using task-related data~\cite{herevisiting}, enhancing supervised fine-tuning with retrieved document~\cite{zhang2024raft}, and incorporating CoT data into fine-tuning~\cite{luong2024reft} can all strengthen the reasoning ability of LLMs by adjusting their parameters.
Recently, the release of OpenAI's o1 has inspired many works~\cite{zhong2024evaluation,zhang2024llama} to explore reasoning techniques similar to o1, emphasizing the importance of scaling CoT and reinforcement learning to strengthen LLM reasoning abilities.
However, all of the aforementioned works primarily focus on open-domain problems such as mathematics, code, and commonsense reasoning, without addressing explicit reasoning in the legal domain. This gap is the focus of our paper, which aims to equip LLMs with the ability to perform explicit syllogistic legal reasoning.

% \vspace{-3mm}

\section{Method}
\subsection{Problem Statement}
We focus on the legal question-answering (QA) task in this paper. Formally, suppose we are given a set of labeled data tuples $\mathcal{T} = \{(x_i, y_i)\}_{i=1}^{|\mathcal{T}|}$, where $|\mathcal{T}|$ is the total number of tuples. For each tuple, $x_i$ represents the $i$-th user's question, and $y_i$ is the manually annotated answer corresponding to the question.

Unlike current Legal QA studies, which directly generate an answer $y_i$ from a given user question $x_i$, we aim to empower LLMs with explicit syllogistic legal reasoning capabilities. Specifically, the objective of our work is to produce $\hat{y}_i$ to $x_i$ by constructing a structured reasoning path $\mathcal{P}_i$. This path includes a major premise $p_i^{\mathrm{major}}$, derived from relevant legal statutes and precedent cases, and a minor premise $p_i^{\mathrm{minor}}$, extracted from the specific facts in $x_i$, which are combined to logically infer $\hat{y}_i$ as the conclusion.

\begin{figure*}
    \centering
    \includegraphics[width=0.95\textwidth]{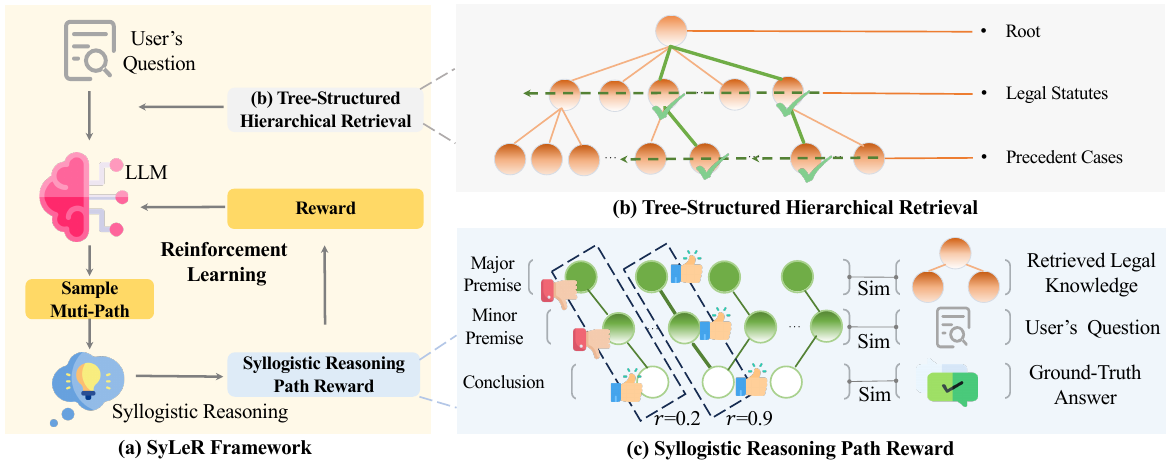}
    \caption{ The architecture of SyLeR. (a) The framework of SyLeR, which includes a tree-based retriever and trains the LLM by sampling multiple syllogistic reasoning paths through reinforcement learning, with rewards based on the reasoning paths. (b) The tree-based retriever, which first retrieves legal statutes and then retrieves corresponding cases under those statutes. (c) The calculation of reasoning path rewards, which is obtained by measuring the similarity between the model-generated major premise, minor premise, and conclusion with the retrieved legal knowledge, the user's question, and the answer.
    }
    \label{fig:framework}
\end{figure*}

\subsection{Overall Framework}
As illustrated in Figure~\ref{fig:framework}, the proposed SyLeR framework consists of two primary components designed to achieve explicit syllogistic legal reasoning.

First, given a user’s legal question $x_i$, SyLeR retrieves the relevant legal foundation to construct the major premise in the syllogistic reasoning process. This is achieved through a hierarchical retrieval mechanism applied to the constructed legal knowledge tree $\mathcal{D}$:
\begin{equation}
\label{eq:retrieval}
\mathcal{K}_i = \mathrm{Retriever}(x_i, \mathcal{D}),
\end{equation}
where $\mathcal{K}_i \subseteq \mathcal{D}$ represents the subset of retrieved legal knowledge, including legal statutes and precedent cases relevant to $x_i$. This retrieval mechanism ensures the interpretive connection between statutes and cases is preserved. The details of the corpus construction and retrieval process are elaborated in \textsection~\ref{sec:tree}.

With the major premise established, SyleR is responsible for deducing the conclusion based on the user’s question, which serves as the minor premise. This step integrates the retrieved knowledge $\mathcal{K}_i$ with the question $x_i$ to produce the final legally coherent  syllogistic reasoning path $\mathcal{P}$:
\begin{equation}
\mathcal{P}_i = \mathrm{LLM}(\mathcal{K}_i, x_i).
\end{equation}
where $\mathcal{P}_i=\{p_i^{\mathrm{major}}, p_i^{\mathrm{minor}}, \hat{y}_i\}$ is the structured syllogistic output including the generated major premise, minor premise, and conclusion.
To enhance reasoning diversity and ensure logical consistency, the framework leverages multiple reasoning paths and a structured reward mechanism, which are detailed in \textsection~\ref{sec:rl}.

\subsection{Tree-Structured Hierarchical Retrieval}\label{sec:tree}
The first step in SyLeR is to retrieve relevant legal knowledge to construct the major premise required for explicit syllogistic legal reasoning. This step involves a hierarchical retrieval mechanism applied to the legal knowledge tree $\mathcal{D}$. As discussed, this tree is designed to preserve the interpretive connections between legal statutes and precedent cases, ensuring a robust and logically sound foundation for reasoning. In this section, we describe how the legal knowledge tree is constructed and how the retrieval mechanism operates.

\textbf{Construction of legal knowledge tree $\mathcal{D}$.}
The legal knowledge corpus $\mathcal{D}$ is organized into a hierarchical tree structure to capture the relationships between general legal principles (legal statutes) and specific applications (precedent cases). As shown in Figure~\ref{fig:framework} (b), the tree consists of three key components:
(1) The root of the tree is a virtual node representing the starting point of retrieval.
(2) The first layer contains nodes representing legal statutes $\mathcal{D}_L$, which provide general legal principles relevant to the user’s question. These statutes form the backbone of the major premise by defining the foundational rules of the legal system.
(3) The second layer contains nodes representing precedent cases $\mathcal{D}_C$, which are linked to their corresponding statutes in $\mathcal{D}_L$. Precedent cases illustrate how the general principles from statutes have been interpreted and applied in specific legal scenarios.
This hierarchical structure ensures that the major premise is comprehensive, combining both general principles and contextualized interpretations. Additionally, the tree structure allows for efficient navigation and retrieval of relevant knowledge.

To construct $\mathcal{D}$, we use the legal corpus provided by~\cite{zhang2024citalaw}, which contains 54,114 statutes and 435,579 cases. Each statute and case is encoded into a dense vector representation using BGE~\cite{bge_embedding}. 
Given the vector representation for each case, we identify the most relevant statute from the corpus based on the cosine similarity of their embeddings, creating an edge between the statute and the case.
To enhance connectivity, statutes with fewer than five connected cases are supplemented by linking additional cases. These additional cases are determined by identifying the most relevant statute and its associated cases based on cosine similarity.
This approach ensures that the tree captures meaningful relationships between statutes and cases while maintaining sufficient connectivity for robust retrieval and reasoning.

\textbf{Hierarchical retrieval.} Given a user question $x_i$, the retrieval process leverages the hierarchical tree structure to retrieve relevant statutes and cases efficiently, as illustrated in Figure~\ref{fig:framework} (b). The retrieval mechanism proceeds in two steps. In the first-layer statue retrieval, the user question $x_i$ is embedded using BGE~\cite{bge_embedding}. We compute the cosine similarity between the question embedding and the embeddings of all statutes in the statute layer. The top-$k$ most similar statutes are selected as retrieval results, which are denoted by $L = \{l_1, l_2, \dots, l_{K_L}\}$, where $K_L$ represents the number of retrieved statutes. In this paper, as the legal question typically relate to a single statute, we set $K_L = 1$. 
In the second-layer precedent cases retrieval,
for each retrieved statute $L_i$, we traverse its associated case nodes in the tree. For each case under $L_i$, we compute the cosine similarity between the question embedding and the case embeddings. We select the top-$k$ most similar cases for each statute, which are denoted by $C_i = \{c_{i1}, c_{i2}, \dots, c_{i{K_C}}\}$,
where $C_i$ represents the set of retrieved cases under statute $L_i$. 
% All retrieved cases across the statutes are aggregated to form the case-layer retrieval results.
To address the input length limitations of LLMs, we set $K_C = 3$, ensuring that only the most relevant cases are included in the retrieved subset.
By combining these two steps, the retrieval mechanism outputs $\mathcal{K}_i$, a subset of legal knowledge containing both statutes and precedents that are logically and semantically connected. This ensures that the major premise for syllogistic reasoning is comprehensive, cohesive, and tailored to the user’s legal question. The retrieved knowledge $\mathcal{K}_i$ is then passed to the next stage of SyLeR for structured reasoning.

\vspace{-2mm}
\subsection{Two-Stage Reinforcement Fine-Tuning}
\label{sec:rl}
With the relevant legal knowledge $\mathcal{K}_i$ retrieved using the tree-structured hierarchical retrieval mechanism, the next step in SyLeR is to enable the LLM to generate explicit syllogistic reasoning paths and produce the legally coherent answer $\hat{y}_i$. This is achieved through a reinforcement fine-tuning process that consists of two stages: a warm-up stage using supervised fine-tuning and a reinforcement learning (RL) stage. 
This fine-tuning process ensures that the model generates logically sound and interpretable reasoning paths that adhere to the major premise $\to$ minor premise $\to$ conclusion structure.

\textbf{Warm-up stage} establishes the LLM’s foundational ability to perform syllogistic legal reasoning. Specifically,
we employ GPT-4o~\cite{hurst2024gpt} to generate 10 syllogistic reasoning paths for each of $(x, y)\in\mathcal{T}$ using the corresponding retrieved documents $\mathcal{K}$ as shown in Eq.~\ref{eq:retrieval}. Each reasoning path $\mathcal{P}$ consists of 
the major premise, derived by summarizing the retrieved statutes and precedent cases in $\mathcal{K}$;
the minor premise, which originates from the specific facts provided in the question $x$;
and the conclusion, corresponding to the ground-truth answer $y$.
Using these reasoning paths, we perform supervised fine-tuning with LoRA~\cite{hu2021lora} to train the LLM to generate responses in the structured format
% of major premise $\to$ minor premise $\to$ conclusion 
as follows:
\begin{equation}
\label{eq:loss_1}
\small
    \mathcal{L}_{\mathrm{FT}} = - \frac{1}{|\mathcal{T}|} \sum_{\mathcal{T}} \mathrm{log}\left(P_{\theta+\theta_{L}}(\mathcal{P}_t|x,\mathcal{K},\mathcal{P}_{<t})\right),
\end{equation}
where $\theta$ and $\theta_{L}$ are the parameters of LLM and LoRA; $\mathcal{P}_t$ and $\mathcal{P}_{<t}$ respectively denote the $t$-th token and tokens before $\mathcal{P}_t$ in the corresponding syllogistic reasoning path. 
This stage provides the model with an initial understanding of syllogistic reasoning and prepares it for further refinement during the reinforcement learning stage.

\textbf{Reinforcement learning.}
Building upon the warm-up stage, we employ the Proximal Policy Optimization (PPO) algorithm~\cite{schulman2017proximal} to refine the model’s reasoning capabilities. The reinforcement learning stage focuses on enabling the LLM to explore multiple reasoning paths for each question, avoiding overfitting to a single reasoning strategy and enhancing the diversity of logical reasoning.
As shown in Figure~\ref{fig:framework} (c), for the response where only the correct answer is provided without a reasonable reasoning path, we assign a lower reward, as the lack of a correct reasoning process can affect the reliability of the reasoning.
During training, LLM samples reasoning paths for a given question $x_i$ by combining the retrieved knowledge $\mathcal{K}_i$ with the question. Each sampled reasoning path is evaluated based on a reward mechanism designed to assess its quality and adherence to the syllogistic structure. PPO optimizes the model by iteratively adjusting the policy to maximize the expected reward while ensuring the updates remain within a stable region to avoid overfitting or catastrophic updates.

\textbf{Reward design for syllogistic reasoning.} 
The reward mechanism plays a crucial role in guiding LLM to generate logically sound syllogistic reasoning paths. For each sampled reasoning path, the reward is computed based on the following criteria:
\begin{align}
\label{eq:reward}
r &= 
\begin{cases}
0 & \text{if not syllogistic,}  \\
\frac{1}{2} \mathrm{sim}(p^\mathrm{major}, l) 
+\frac{1}{2}\sum\limits_{i=1}^{3} \frac{1}{3}\mathrm{sim}(p^\mathrm{major}, c_i)\\
\quad\quad  +\mathrm{sim}(p^{\mathrm{minor}}, x) + \mathrm{rouge}_{\mathrm{sum}}(\hat{y}, y) & \text{otherwise,}
\end{cases}
\end{align}
$\mathrm{sim}(\cdot)$ represents the cosine similarity computed using Sentence-BERT embeddings~\cite{reimers2019sentence}, while $\mathrm{rouge}_{\mathrm{sum}}(\cdot)$ denotes the exponential form of combined Rouge-1, Rouge-2, and Rouge-L scores. The terms in Eq.~\ref{eq:reward} are defined as follows:
\begin{itemize}[leftmargin=*]
\item $\mathrm{sim}(p^\mathrm{major}, l)$ measures the alignment between the generated major premise ($p^\mathrm{major}$) and the retrieved statute ($l$).
\item $\sum\limits_{i=1}^{3} \frac{1}{3}\mathrm{sim}(p^\mathrm{major}, c_i)$ averages the similarity between the major premise and the three retrieved precedent cases $\{c_1, c_2, c_3\}$.
\item $\mathrm{sim}(p^\mathrm{minor}, x)$ evaluates how well the generated minor premise ($p^\mathrm{minor}$) aligns with the question ($x$).
\item $\mathrm{rouge}_{\mathrm{sum}}(\hat{y}, y)$ measures the similarity between the generated conclusion ($\hat{y}$) and the ground-truth answer ($y$) using ROUGE metrics.
\end{itemize}
The reward $r$ is only assigned at the terminal state, i.e., when the generation is completed. All intermediate actions in non-terminal states receive a reward of zero. To further enforce the adherence to syllogistic reasoning, any invalid output—such as one that deviates from the expected major premise $\to$ minor premise $\to$ conclusion structure—is penalized with a reward of zero. 

Following~\cite{zheng2023secrets,luong2024reft}, the total reward also incorporates a KL divergence~\cite{kullback1951information} term between the RL model and the warm-up model. This regularization stabilizes training by penalizing the RL policy for diverging excessively from the warm-up policy. The final reward is defined as:
\begin{equation}
r_\mathrm{final} = r - \beta \mathrm{KL}\left(\pi_\theta(\cdot | s_t), \pi_{\theta_\mathrm{SFT}}(\cdot | s_t)\right),
\end{equation}
where $\pi_\theta$ and $\pi_{\theta_\mathrm{SFT}}$ represent the RL policy and the warm-up stage policy, respectively, and $\beta$ is a scaling coefficient that controls the contribution of the KL divergence to the overall reward.
By combining task-specific rewards with a KL regularization term, this design encourages the model to explore diverse reasoning paths while maintaining stability and alignment with the initial supervised policy.

\section{Experiment}
In this section, we conduct experiments to answer the following research questions:
\textbf{RQ1:} How does SyLeR perform compared to existing legal LLMs and open-domain LLM baselines in legal layperson and practitioner datasets? \textbf{RQ2:} What is the impact of the three modules in SyLeR on the model's performance?  \textbf{RQ3:} How does SyLeR perform across different domains? \textbf{RQ4:} How robust is SyLeR across different languages? \textbf{RQ5:} Can SyLeR be adapted to various LLM backbones? \textbf{RQ6:} Are the responses generated by SyLeR truly trustworthy?

\subsection{Experimental Setup}
\subsubsection{Dataset.}
We conducted experiments on three legal QA datasets: two Chinese datasets, Practitioner and Layperson, targeting different user groups, and one French dataset, LLeQA.

\textbf{Practitioner}~\cite{li2024lexevalcomprehensivechineselegal}: This dataset is derived from LexEval, which collects questions and answers from the subjective sections of the National Unified Legal Professional Qualification Examination. It represents questions from legal professionals, including detailed case descriptions that require LLMs to thoroughly understand the case context and generate answers based on relevant legal knowledge.

\textbf{Layperson}~\cite{fei2023lawbench}: This dataset is sourced from LawBench, which compiles user questions and responses from the Hualv website\footnote{www.66law.com}, where non-legal professionals seek advice from lawyers and legal practitioners. The questions are shorter and more conversational in nature, requiring LLMs to leverage legal knowledge for appropriate responses.
For the corpus associated with these two datasets, we utilized a knowledge base provided by~\cite{zhang2024citalaw}, which includes legal statutes and precedent cases.

\textbf{LLeQA}~\cite{louis2024interpretable}: This is a French dataset with expert-annotated legal questions covering diverse topics such as housing, family, and employment. The questions often require longer answers, demanding that LLMs refer to relevant legal statutes to provide detailed responses. For the corpus of this dataset, we used~\cite{louis2024interpretable}, which contains 27,942 French legal statutes.

\begin{table}[t]
\centering
\caption{
The statistical analysis of the datasets used in this paper. \#Train and \#Test represent the number of data points in the training and testing sets, respectively. Len$_Q$ and Len$_A$ refer to the average question length and answer length. 
}
\resizebox{0.35\textwidth}{!}{
\begin{tabular}{l|cccc}
    \toprule
    Dataset &\#Train &\# Test & Len$_Q$ & Len$_A$  \\
    \hline
    Layperson &100 &500 &57.62&107.40\\ 
    \hline
    Practitioner&100 &500 &618.96&193.46 \\
    \hline
    LLeQA &1472&195&15.39&305.92\\
    \bottomrule
\end{tabular}
}
\label{tab:data_main}
 % \vspace{-0.3cm}
\end{table}

Statistical analyses of the three datasets are shown in Table~\ref{tab:data_main}. We fine-tuned LLMs using the training subsets of these datasets and evaluated performance on their respective test sets.

\subsubsection{Evaluation Metrics.}
We employed widely used metrics to evaluate the similarity between model-generated responses and reference answers, including Rouge~\cite{lin2004rouge} (Rouge-1, Rouge-2, and Rouge-L), which assess recall; BLEU~\cite{papineni2002bleu}, which evaluates precision; and BERTScore (BERT)~\cite{zhang2019bertscore}, which measures semantic similarity by leveraging contextual embeddings.

\begin{table*}[t]
    \small
\caption{Performance comparison between SyLeR and the baseline on the Layperson and Practitioner datasets, with the best performance highlighted in bold and the second-best method underlined.}
    \label{tab:main exp}
\centering
    \begin{tabular}{cc|ccccc|ccccc}    
        \toprule &
        &\multicolumn{5}{c|}{Practitioner}
        &\multicolumn{5}{c}{Layperson}
        \\
         \textbf{Category} &\textbf{Models}  & \textbf{Rouge-1} & \textbf{Rouge-2}& \textbf{Rouge-L} & \textbf{BLEU}& \textbf{BERT} & \textbf{Rouge-1} & \textbf{Rouge-2}& \textbf{Rouge-L} & \textbf{BLEU}& \textbf{BERT}\\

        \midrule
        % \multicolumn{7}{c}{Legal LLM} \\
        % \midrule
        \multicolumn{1}{c}{\multirow {7}{*}{\shortstack{Legal LLM}}}  
        &zhihai & 21.89 & 6.19 & 16.30 & 5.13 & 64.56 & 14.65 & 1.72 & 11.00 & 2.74 & 59.64 \\
        &fuzi.mingcha & 28.94 & 9.51 & 22.61 & 8.39 & 66.94 & 23.42 & 4.83 & 16.50 & 5.99 & 65.72\\
        &DISC-LawLLM & 23.07 & 7.28 & 18.21 & 6.14 & 62.55 & 13.23 & 2.12 & 9.80 & 3.12 & 58.00\\
        &LawGPT\_zh & 29.50 & 9.14 & 22.93 & 7.85 & 67.57 & 22.67 & 4.12 & 16.12 & 4.85 & 65.02 \\
        &Tailing & 27.93 & 9.58 & 23.50 & 5.37 & 65.89 & 21.09 & 4.09 & 16.65 & 4.26 & 62.94 \\
        &LexiLaw & 29.90 & 9.27 & 24.07 & 7.09 & 65.71 & \underline{24.62} & 5.03 & 18.27 & \underline{6.25} & \underline{65.71}\\
        &Hanfei & 33.20 & 12.39 & 27.06 & 9.68 & 68.74 & 23.94 & 4.74 & \underline{18.51} & 5.63 & 65.31\\
        \midrule
        \multicolumn{1}{c}{\multirow {8}{*}{\shortstack{Qwen2\\(Qwen2-7B-Instruct)}}}  
        &Standard RAG & 29.97 & 10.19 & 22.82 & 9.08 & 68.77 & 14.34 & 2.62 & 10.54 & 2.58 & 58.72  \\
&Selective-context & 30.02 & 10.76 & 23.43 & 9.59 & 66.61 & 9.89 & 1.13 & 7.63 & 1.31 & 52.27   \\
&LongLLMLingua & 32.15 & 10.65 & 24.62 & 10.31 & 69.52 & 18.78 & 3.33 & 13.80 & 4.01 & 62.54  \\
&Iter-Retgen & 29.65 & 9.88 & 22.61 & 8.69 & 68.47 & 13.87 & 2.47 & 10.12 & 3.17 & 58.88 \\
\cdashline{2-12}
&Naive SFT & 32.79 & \underline{13.43} & 26.61 & 10.33 & 69.70 & 22.42 & \underline{5.92} & 17.60 & 3.73 & 64.65   \\
&RAG-SFT & \underline{33.70} & 13.01 & 26.39 & \underline{11.41} & \underline{69.90} & 19.90 & 4.07 & 14.16 & 5.58 & 64.30   \\
&CoT-SFT & 32.58 & 11.64 & \underline{27.09} & 7.87 & 67.76 & 21.65 & 5.06 & 17.48 & 4.80 & 64.75   \\
% tree\_cot & 34.24 & 13.03 & 27.18 & 10.70 & 69.48 & 25.79 & 6.44 & 19.00 & 7.27 & 66.76   \\
        % \midrule
        \cdashline{2-12}
&SyLeR (Ours) & \textbf{36.01} & \textbf{13.83} & \textbf{29.56} & \textbf{11.81} & \textbf{70.55} & \textbf{27.19} & \textbf{7.05} & \textbf{21.11} & \textbf{9.05} & \textbf{67.90} \\

        \bottomrule
    \end{tabular}
\end{table*}

\subsubsection{Baselines}
We compare two types of LLMs: legal-specific LLMs and open-domain LLMs.
Legal LLMs include \textbf{zhihai}~\cite{wisdomInterrogatory}, \textbf{fuzi.mingcha}~\cite{sdu_fuzi_mingcha}, \textbf{DISC-LawLLM}~\cite{yue2023disclawllm}, \textbf{LawGPT\_zh}~\cite{LAWGPT-zh}, \textbf{Tailing}\footnote{https://github.com/DUTIR-LegalIntelligence/Tailing}, \textbf{LexiLaw}\footnote{https://github.com/CSHaitao/LexiLaw}, \textbf{HanFei}~\cite{HanFei}. These models have been fine-tuned on extensive datasets from the legal domain, enabling them to handle legal tasks.

We used Qwen2~\cite{qwen2} as the representative open-domain LLM for our experiments.
For open-domain LLMs, 
in addition to the \textbf{standard RAG}~\cite{lewis2020retrieval,gao2023retrieval} approach, where the LLM generates responses based on retrieved laws and related cases, we compared prompt-based methods designed to handle the long document lengths common in legal scenarios:
\textbf{Selective-Context}~\cite{li2023compressing} removes redundant information from the provided context using self-information~\cite{shannon1948mathematical}.
\textbf{LongLLMLingua}~\cite{jiang2023longllmlingua} employs a coarse-to-fine, question-aware compression strategy to condense lengthy documents.
To address the limitation of single-pass retrieval, which may fail to provide all the knowledge needed for legal questions, we also compared the iterative retrieval method \textbf{Iter-RetGen}~\cite{shao2023enhancing}.

For fine-tuning methods, we compared the following approaches:
\textbf{Naive SFT}~\cite{herevisiting}: A straightforward supervised fine-tuning approach using question-answer pairs for training.
\textbf{RAG-SFT}~\cite{zhang2024raft}: Fine-tuning the LLM using questions combined with retrieved documents as input and answers as output.
\textbf{CoT-SFT}~\cite{luong2024reft}: This method uses GPT-4o to generate direct reasoning paths based on the question, retrieved legal documents and answer. During fine-tuning, the question and documents are used as input, while the reasoning path and conclusion are used as output to train the LLM.

\subsubsection{Implementation Details.}
For all models in the main experiments, we use BGE (specifically, bge-base-zh-v1.5)~\cite{bge_embedding} to retrieve one legal statute and three precedent cases. We use the training dataset for fine-tuning and reinforcement learning training.
For our method, following~\cite{luong2024reft}, during the warm-up stage, we fine-tune for 3 epochs. During the RL stage, we train for 10 epochs. For fine-tuning baselines, we train for 10 epochs to ensure full convergence. In the fine-tuning and warm-up stage, we set the maximum input length to 4096, the learning rate to 5e-5, and the batch size to 16. For both the fine-tuning and RL stages, we use LoRA~\cite{hu2021lora} for efficient training of the LLM, setting LoRA parameters with a rank of 8 and an $\alpha$ of 16.
For the RL stage, we use PPO~\cite{schulman2017proximal} for training, with the learning rate set to 1e-5, $\beta$ set to 0.02, the clip range set to 0.2, and the batch size set to 16. 
During LLM generation, we set do\_sample to false to enhance the reproducibility of the results.
For the calculation of BERTScore, we use bert-base-chinese~\cite{devlin2018bert} for the Layperson and Practitioner datasets, and bert-base-multilingual-cased for the LLeQA dataset.
All experiments are conducted on A6000 GPUs.

\subsection{Overall Performance (RQ1)}
Table~\ref{tab:main exp} presents the results of SyLeR and the baselines on the Layperson and Practitioner datasets. We can draw the following conclusions:

\textbf{SyLeR has the best overall performance.}
It can be seen that SyLeR achieves the best performance across all metrics on both datasets, which demonstrates the effectiveness of SyLeR. SyLeR enhances the retrieval of legal statutes and precedent cases with stronger relevance through a tree-structured hierarchical retrieval mechanism, providing the LLM with more structured legal knowledge. This structured knowledge helps the LLM better understand and apply relevant legal statutes and precedent cases, leading to improved reasoning ability.
In addition, SyLeR uses reinforcement learning to sample various syllogistic reasoning paths and assigns reward signals based on the major premise, minor premise, and conclusion during the reasoning process, effectively training the LLM. 
% This multi-path training method enables the LLM to learn from diverse reasoning angles, further enhancing its syllogistic reasoning capabilities. 
As a result, the LLM’s responses become more accurate.
Compared to legal LLMs, which have been fine-tuned with a large amount of legal knowledge, SyLeR outperforms them due to they primarily rely on implicit reasoning to obtain answers, which results in inferior performance.
Compared to baseline methods in open-domain, these methods may rely solely on a single reasoning path, such as SFT. In contrast, SyLeR offers various reasoning paths, which leads to improved reasoning accuracy in the final model outputs.

\textbf{Legal LLMs performer well.}
Although the base LLM used for the legal LLM is an earlier version of open-domain LLM and performs worse than Qwen2 on various tasks, the legal LLM still demonstrates strong performance. This is because it has been fine-tuned on a large amount of legal-related datasets, embedding a significant amount of legal knowledge. Even without adding any strategies, such as compressing the retrieved documents, it performs second only to our proposed SyLeR method on certain metrics, such as the Lexilaw task on the Layperson dataset.
This result highlights the importance of legal knowledge in enhancing the performance of LLMs for legal tasks. It also provides insight for our research, suggesting that future exploration could focus on how to combine the SyLeR method with more extensive and richer legal knowledge, not only maintaining the depth and breadth of the knowledge but also further improving the model's reasoning capabilities for complex legal tasks. 

\textbf{The benefits of fine-tuning.}
As observed, methods that enhance LLM reasoning through fine-tuning have outperformed methods that enhance reasoning based on prompts across both datasets. This result suggests that solely relying on improving the prompts to boost LLM reasoning capabilities is insufficient for legal reasoning tasks. 
This is because legal reasoning is significantly different from traditional reasoning scenarios. Legal problems typically involve more complex legal concepts, rules, and cases, requiring more specialized knowledge. After retrieving relevant legal knowledge, effectively using it to solve specific legal issues is not an easy task. 
Additionally, prompt-based methods do not guarantee that the LLM can properly use the retrieved legal knowledge to provide a reasonable answer, especially when handling complex legal cases. 
However, the amount of training data in legal scenarios is relatively limited, making it a key challenge to explore and provide efficient methods to equip LLMs with legal reasoning capabilities. 
This is the focus of this paper. 
SyLeR offers an efficient solution by exploring explicit syllogistic reasoning paths through reinforcement learning, effectively utilizing legal knowledge and enhancing the model’s performance in complex legal reasoning tasks. 
Moreover, SyLeR enables the LLM to generate responses that include syllogistic reasoning, and provides a clear reasoning path that makes the model-generated responses more trustworthy and explainable.

\subsection{Ablation Study (RQ2)}\label{sec:ablation}

\begin{table}
\centering
    \caption{Ablation experiments of SyLeR on the Layperson and Practitioner datasets. w/o Path Reward indicates the removal of the reward that ensures the reasoning path. w/o Tree Retrieve means replacing the tree retrieve with a regular retrieve. w/o RL refers to the removal of the RL training that explores multiple syllogistic reasoning paths. The best performance is indicated in bold.}
{
\renewcommand{\arraystretch}{1.1}
\resizebox{0.98\columnwidth}{!}{
\begin{tabular}{l ccc cc}
\toprule
Model
   & \textbf{Rouge-1} & \textbf{Rouge-2}& \textbf{Rouge-L} & \textbf{BLEU}& \textbf{BERT} \\
    \midrule
    \multicolumn{6}{c}{\textbf{Practitioner}} \\
    \midrule
    SyLeR &   \textbf{36.01} & \textbf{13.83} & \textbf{29.56} & \textbf{11.81} & \textbf{70.55}     \\
    \hdashline  
    w/o Path reward     & 34.60 & 13.06 & 27.52 & 11.45 & 69.83          \\
    w/o Tree Retrieve   & 35.46 & 13.06 & 29.03 & 9.03 & 69.33      \\
    w/o RL    & 34.24 & 13.03 & 27.18 & 10.70 & 69.48     \\    
    \midrule
    \midrule
    \multicolumn{6}{c}{\textbf{Layperson}} \\
    \midrule
    SyLeR  & \textbf{27.19} & \textbf{7.05} & \textbf{21.11} & \textbf{9.05} & \textbf{67.90}          \\
    \hdashline  
      w/o Path reward  & 26.50 & 6.25 & 19.76 & 9.00 & 67.77     \\
      w/o Tree Retrieve  & 26.56 & 7.01 & 20.32 & 7.48 & 66.80    \\  
    w/o RL& 25.79 & 6.44 & 19.00 & 7.27 & 66.76\\
    \bottomrule
\end{tabular}}
}%
    \label{tab:ablation}
\end{table}
SyLeR primarily consists of three key components: reasoning path validation, tree-based retrieval of legal statutes and precedent cases, and RL-based exploration of multiple reasoning paths. To explore how these three components impact the performance of SyLeR, we conducted ablation experiments on the Layperson and Practitioner datasets by removing each component from the overall method. The results, as shown in Table~\ref{tab:ablation}, are analyzed in detail below:

\textbf{w/o Path Reward}: We removed the reward that ensures the reasoning path is valid, specifically eliminating the rewards for the major and minor premises while keeping only the reward for the conclusion during RL training. The results show a decline in both datasets, indicating the importance of the path reward. Ensuring that the major premise aligns with the retrieved legal statutes and precedent cases and that the minor premise aligns with the user’s question, helps stabilize the model’s reasoning process and guides the model toward more accurate answers.

\textbf{w/o Tree Retrieve}: In this experiment, we replaced the tree-based retrieval for retrieving legal statutes and precedent cases with individual searches for each component (i.e., separately retrieving legal statutes and precedent cases). Other training settings remained unchanged. The results show a noticeable decline, emphasizing the importance of tree-based retrieval. The tree-based retrieval provides a more systematic and structured set of legal knowledge, helping the LLM better understand and solve legal problems by retrieving more relevant documents. This approach also aids in enhancing the model’s reasoning abilities.

\textbf{w/o RL}: SyLeR samples multiple syllogistic reasoning paths during the RL training stage to fully train the LLM. In this ablation, we removed this stage and only tested the model trained after the warm-up stage. This allows us to assess the role of RL in SyLeR. The results show a significant drop in performance without RL, indicating the importance of sampling multiple reasoning paths. Simply relying on the warm-up stage’s SFT does not equip the LLM with strong reasoning capabilities. RL is crucial in enhancing the model’s ability to generate diverse reasoning paths, improving its overall performance in legal reasoning tasks.

These ablation experiments demonstrate that each component plays a vital role in SyLeR, and removing any of them leads to a noticeable decrease in performance, underscoring their importance in enhancing the model's reasoning abilities.

\subsection{Cross Domain Experiment (RQ3)}\label{sec:exp cross}
\begin{figure}[t]
    \centering
        \subfigure[Layperson to Practitioner.]
    {
    \includegraphics[width=0.47 \linewidth]{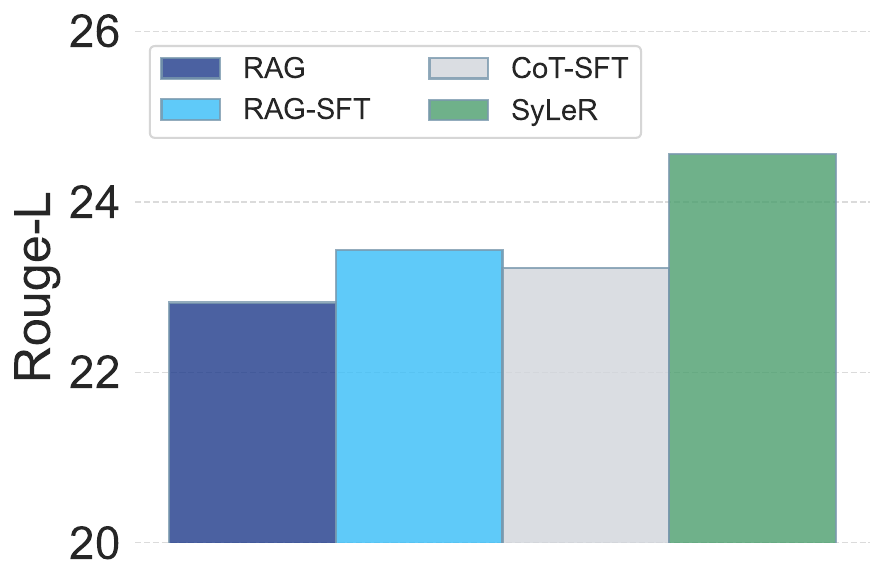}
    \label{fig:lay to pra}
    }
    \subfigure[Practitioner to Layperson.]
    {
    \includegraphics[width=0.47 \linewidth]{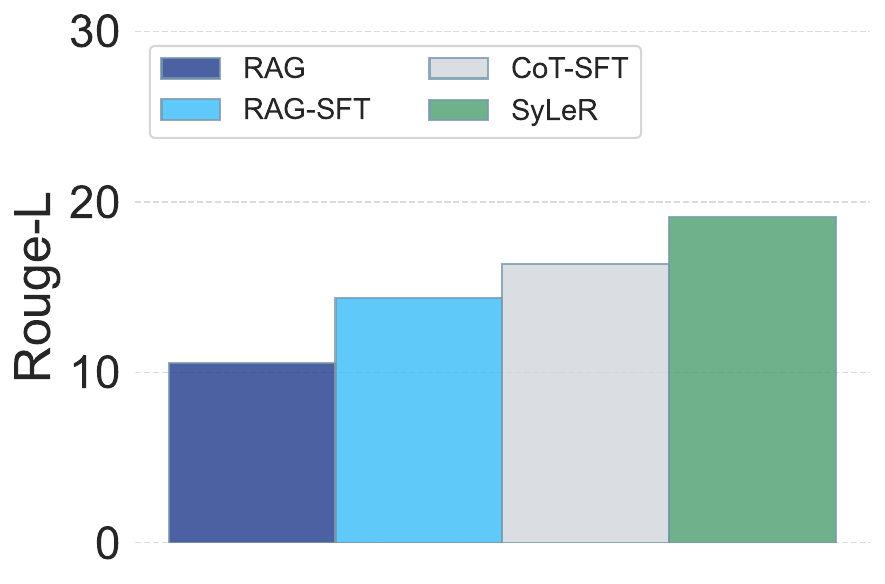}
    \label{fig:pra to lay}
    }
    \caption{
Results of the cross-domain experiments. (a) Applying the model trained on the Layperson dataset to test on the Practitioner dataset. (b) Applying the model trained on the Practitioner dataset to test on the Layperson dataset.
    }
\label{fig:cross exp}
\end{figure}

\begin{table}
\centering
    \caption{Comparison of SyLeR and baselines on the French QA dataset LLeQA. The best performance is indicated in bold.}
{
\renewcommand{\arraystretch}{1.1}
\resizebox{0.9\columnwidth}{!}{
\begin{tabular}{l ccc cc}
\toprule
Model
   & \textbf{Rouge-1} & \textbf{Rouge-2}& \textbf{Rouge-L} & \textbf{BLEU}& \textbf{BERT} \\
    \midrule
    Standard RAG & 23.40 & 4.58 & 21.50 & 2.15 & 68.49    \\
    Naive SFT & 27.81 & 8.05 & 26.33 & 2.86 & 69.62\\
    RAG-SFT & 27.75 & 8.21 & 26.33 & 3.09 & 69.30\\
    SyLeR & \textbf{29.04} & \textbf{8.54} & \textbf{27.49} & \textbf{3.61} & \textbf{70.16}\\
    \bottomrule
\end{tabular}}
}%
    \label{tab:fayu}
\end{table}

The cross-domain performance is crucial for legal tasks, where legal knowledge and reasoning skills must be applied across a variety of case types and user profiles. 
The main experiment has validated the effectiveness of our method within the target domain. To further assess the generalization and robustness of SyLeR, we conduct cross-domain experiments in this section. Specifically, we apply the model trained on the Layperson dataset to the Practitioner dataset and vice versa, testing how well the model performs when transferred across domains.

As shown in Figure~\ref{fig:cross exp}, the results indicate that SyLeR achieves optimal performance on both datasets in the cross-domain setting. This demonstrates the strong generalization ability of our method, suggesting that SyLeR not only performs well within a specific domain but can also effectively transfer its reasoning capabilities to a different but related domain.
Furthermore, the results indicate that SyLeR’s tree retrieval of legal statutes and precedent cases, along with its two-stage reinforcement fine-tuning, contribute to its robustness in adapting to different contexts, reinforcing the model’s utility in diverse legal scenarios.
This confirms that SyLeR’s methodology is not domain-specific but has the potential to be applied to a wide range of legal reasoning tasks, further enhancing its practical value in real-world applications.

\subsection{Robustness across Various Languages (RQ4)}\label{sec:ablation}

\begin{figure}
    \centering
\includegraphics[width=0.98\linewidth]{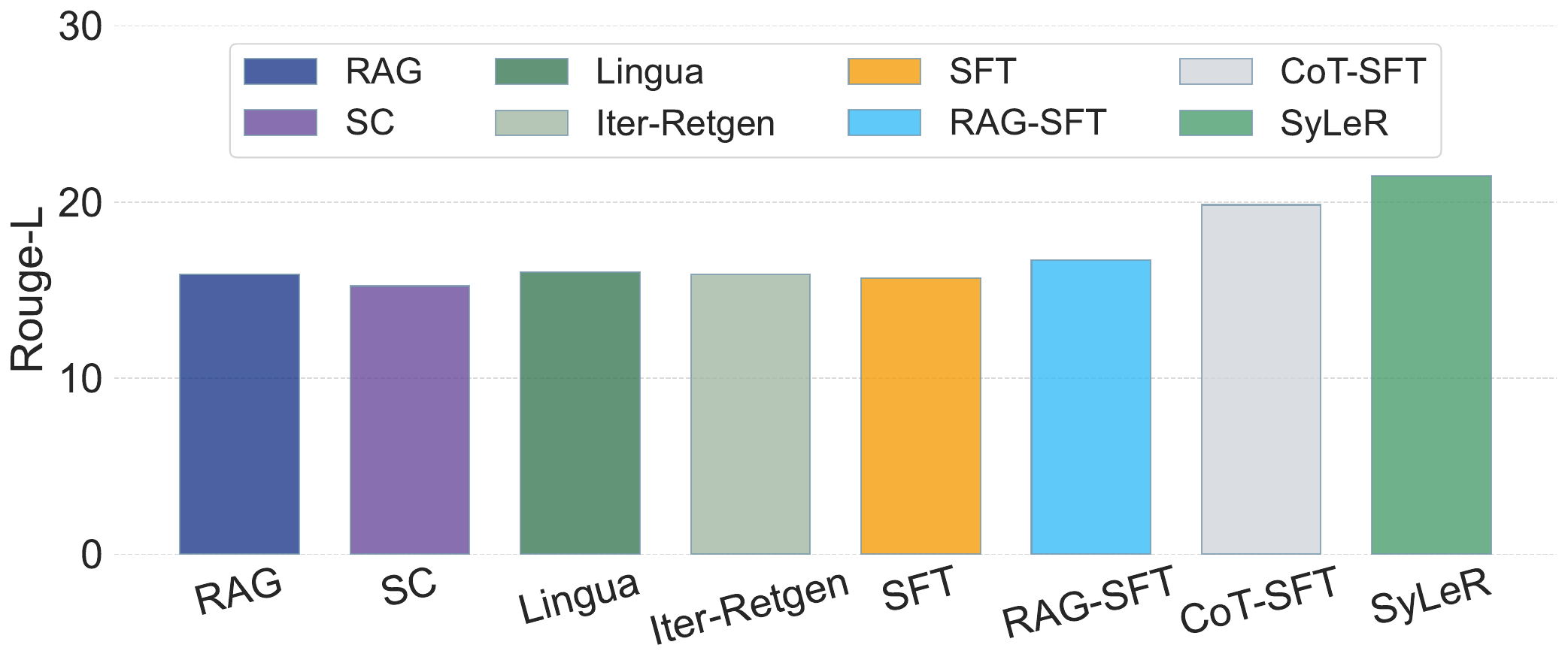}
    \caption{
Comparison of the performance between SyLeR and the baselines on the Layperson dataset when the LLM backbone is Llama3. SC is short for Selective-context method. Lingua is short for LongLLMLingua.
            }
    \label{fig:llama res}
    % \vspace{-3mm} 
\end{figure}

\begin{table}
\centering
    \caption{Human evaluation of the response performance, with each item scored from 1 to 5, with 5 being the highest. The best performance is indicated in bold.}
{
\renewcommand{\arraystretch}{1.1}
\resizebox{0.9\columnwidth}{!}{
\begin{tabular}{l ccc cc}
\toprule
Model
   & \textbf{Correct} & \textbf{Logic}& \textbf{Explain} & \textbf{Trust}& \textbf{Average} \\
    \midrule
    Naive SFT &4.18	&3.76	&3.37	&3.69	&3.75   \\
    RAG-SFT&4.21	&3.97	&3.83	&3.88	&3.97\\
    CoT-SFT &4.41	&4.26	&4.12	&4.23	&4.26\\
    SyLeR &\textbf{4.65}	&\textbf{4.59}	&\textbf{4.66}	&\textbf{4.72}	&\textbf{4.66}\\
    \bottomrule
\end{tabular}}
}%
    \label{tab:qualitative}
\end{table}

\begin{figure*}
    \centering
\includegraphics[width=0.9\linewidth]{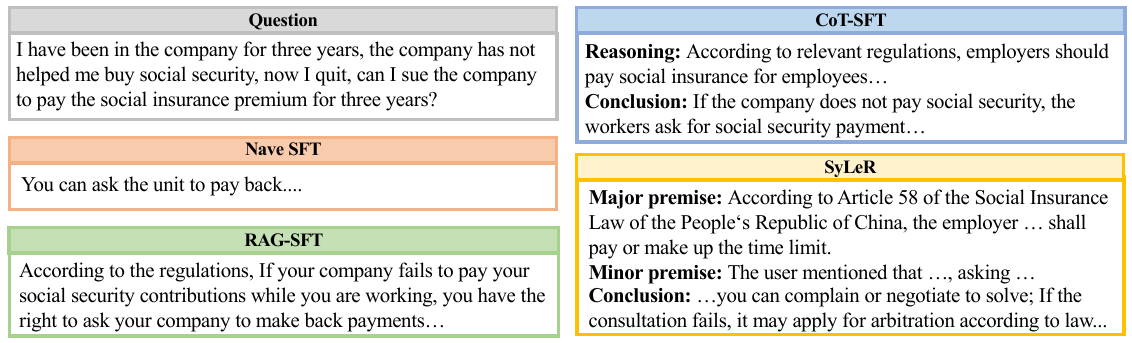}
    \caption{
An example of the responses from SyLeR and baselines.
            }
    \label{fig:case study}
    % \vspace{-3mm}
\end{figure*}

The previous experiments mainly focused on evaluating the performance of our method on Chinese legal tasks. In this section, we explore the performance of SyLeR on the French dataset LLeQA. The LLeQA dataset's document corpus only includes legal statutes, so in this experiment, we omitted the tree-based retrieve module in SyLeR and replaced the retriever with BM25~\cite{robertson1995okapi} to retrieve the most relevant legal statute for the given question.
Given that in LLeQA, typically only one related legal statute is needed for a given question to supplement the LLM's knowledge, we retrieve only the most relevant legal statute to enhance the LLM's legal knowledge in this experiment. 
We compare SyLeR with several other methods, including standard RAG, Naive SFT, and RAG-SFT to assess the performance of different approaches on the French legal QA task.

The experimental results are shown in Table~\ref{tab:fayu}. It can be observed that our method still achieves the best performance on the LLeQA dataset. This indicates the effectiveness of SyLeR. Despite the differences in task types between the LLeQA dataset and the Chinese dataset, SyLeR demonstrates exceptional performance on the French legal task, proving its cross-lingual generalization capability. Moreover, even after omitting the tree-based retrieve module and using BM25 for retrieval, our model still maintains excellent performance. This suggests that SyLeR performs well not only in complex tree-search retrieval tasks but also with simpler retrieval strategies.
Through the comparison with baseline methods, we can see that although these baseline methods can provide some degree of reasoning support, SyLeR's multi-path reasoning enables it to generate more precise and reliable reasoning when handling French legal tasks.

\subsection{Different LLM Backbone (RQ5)}\label{sec:ablation}

In the main experiment, we used Qwen, an LLM with strong performance on Chinese tasks, as the representative model. To further test the performance of our method on other LLMs, in this section, we replace the LLM backbone with Llama3~\cite{llama3modelcard} (Llama3-8B-Instruct) and compare SyLeR with standard RAG, methods for enhancing the reasoning capabilities of LLMs based on prompting and fine-tuning in Layperson dataset.
As shown in Figure~\ref{fig:llama res}, SyLeR still achieves optimal performance, highlighting its effectiveness across different LLM architectures. This result demonstrates that SyLeR is not limited to a specific LLM, and it can be successfully applied to a wide range of LLMs, including those with different underlying architectures like Llama3. The consistent performance across models further validates the robustness of SyLeR in enhancing reasoning capabilities, irrespective of the LLM backbone. It also underscores the importance of our explicit reasoning approach in improving the model’s ability to handle complex legal tasks, even when the backbone LLM may differ in structure.

\subsection{Trustworthiness Verficiation (RQ6)}\label{sec:qualitative}

To explore the trustworthiness of the model's responses, we conducted an evaluation of the quality of the model-generated replies through human judgment in this section. We randomly selected 50 questions from the Layperson dataset and tested the following models: SyLeR, Naive SFT, RAG-SFT, and CoT-SFT. We hired three graduate students majoring in Chinese law to perform the evaluation. All of them have received no less than four years of legal education and have practical legal experience, making them familiar with the questions in the Layperson dataset. After anonymizing the methods, we provided them with multiple 4-tuples (question, retrieved document, true answer, model's response) and asked them to score each response from 1 to 5 on the following aspects, with 5 being the highest:
1) Correctness (Correct): Does the model's response fully address the key points of the question and closely match the true answer?
2) Logicality (Logic): Does the model’s response exhibit a clear and reasonable reasoning process, with no contradictions or gaps, and align with legal reasoning?
3) Explainability (Explain): Does the model's response reasonably apply the content of the retrieved document to provide an explanation? Does it help the user understand how the conclusion was reached?
4) Trustworthiness (Trust): Does the model’s response inspire enough trust, including the use of legal knowledge and cases to demonstrate a high level of professionalism and leave the user with a sense of trustworthiness?
We averaged the scores from the evaluators, and the results are shown in Table~\ref{tab:qualitative}. As can be seen, SyLeR achieved the best performance in all four aspects, indicating that SyLeR not only provides accurate and reasonable answers when handling legal questions but also excels in the explainability of the reasoning process and the trustworthiness of the legal knowledge. SyLeR more effectively combines the retrieved legal documents with the questions, demonstrating stronger reasoning ability and professionalism.

Furthermore, Figure~\ref{fig:case study} shows an example of the model’s response. As seen, SyLeR starts with the legal statutes and performs reasonable legal reasoning in response to the user’s question, integrating relevant legal knowledge to generate an answer. Through a clear reasoning process and accurate citation of the legal statutes, SyLeR not only provides an answer that aligns with legal logic but also enhances the user's trust in the response.

\vspace{-2mm}
\section{Conclusion}

In this paper, we introduced SyLeR, a novel framework designed to equip LLMs with the capability to perform explicit syllogistic legal reasoning. By integrating a tree-structured hierarchical retrieval mechanism, SyLeR effectively combines legal statutes and precedent cases to construct comprehensive and logically connected major premises. 
Through a two-stage fine-tuning process, SyLeR progressively enhances the model’s reasoning capabilities. The first stage, supervised warm-up, establishes a foundational understanding of syllogistic reasoning by training the model to generate structured reasoning paths. The second stage leverages reinforcement learning with a structure-aware reward mechanism to refine the model’s outputs further. This mechanism not only encourages the exploration of diverse reasoning paths but also ensures logical consistency and strict adherence to the syllogistic reasoning format.
Extensive experiments demonstrate that SyLeR consistently improves response accuracy while delivering and trustworthy outputs across different user groups, languages, and LLM backbones. Cross-domain evaluations further validate SyLeR’s adaptability and robustness in diverse legal scenarios, highlighting its potential for real-world legal applications.

% \begin{acks}
% \end{acks}

\bibliographystyle{ACM-Reference-Format}
\bibliography{sample-base}


\begin{thebibliography}{44}

%%% the Web address.

\ifx \showCODEN    \undefined \def \showCODEN     #1{\unskip}     \fi
\ifx \showDOI      \undefined \def \showDOI       #1{#1}\fi
\ifx \showISBNx    \undefined \def \showISBNx     #1{\unskip}     \fi
\ifx \showISBNxiii \undefined \def \showISBNxiii  #1{\unskip}     \fi
\ifx \showISSN     \undefined \def \showISSN      #1{\unskip}     \fi
\ifx \showLCCN     \undefined \def \showLCCN      #1{\unskip}     \fi
\ifx \shownote     \undefined \def \shownote      #1{#1}          \fi
\ifx \showarticletitle \undefined \def \showarticletitle #1{#1}   \fi
\ifx \showURL      \undefined \def \showURL       {\relax}        \fi
% invisible to TeX
\providecommand\bibfield[2]{#2}
\providecommand\bibinfo[2]{#2}
\providecommand\natexlab[1]{#1}
\providecommand\showeprint[2][]{arXiv:#2}

\bibitem[AI@Meta(2024)]%
        {llama3modelcard}
\bibfield{author}{\bibinfo{person}{AI@Meta}.} \bibinfo{year}{2024}\natexlab{}.
\newblock \showarticletitle{Llama 3 Model Card}.
\newblock  (\bibinfo{year}{2024}).
\newblock
\urldef\tempurl%
\url{https://github.com/meta-llama/llama3/blob/main/MODEL_CARD.md}
\showURL{%
\tempurl}


\bibitem[Choi et~al\mbox{.}(2021)]%
        {choi2021chatgpt}
\bibfield{author}{\bibinfo{person}{Jonathan~H Choi}, \bibinfo{person}{Kristin~E Hickman}, \bibinfo{person}{Amy~B Monahan}, {and} \bibinfo{person}{Daniel Schwarcz}.} \bibinfo{year}{2021}\natexlab{}.
\newblock \showarticletitle{ChatGPT goes to law school}.
\newblock \bibinfo{journal}{\emph{J. Legal Educ.}}  \bibinfo{volume}{71} (\bibinfo{year}{2021}), \bibinfo{pages}{387}.
\newblock


\bibitem[Deng et~al\mbox{.}(2023)]%
        {deng2023syllogistic}
\bibfield{author}{\bibinfo{person}{Wentao Deng}, \bibinfo{person}{Jiahuan Pei}, \bibinfo{person}{Keyi Kong}, \bibinfo{person}{Zhe Chen}, \bibinfo{person}{Furu Wei}, \bibinfo{person}{Yujun Li}, \bibinfo{person}{Zhaochun Ren}, \bibinfo{person}{Zhumin Chen}, {and} \bibinfo{person}{Pengjie Ren}.} \bibinfo{year}{2023}\natexlab{}.
\newblock \showarticletitle{Syllogistic reasoning for legal judgment analysis}. In \bibinfo{booktitle}{\emph{Proceedings of the 2023 Conference on Empirical Methods in Natural Language Processing}}. \bibinfo{pages}{13997--14009}.
\newblock


\bibitem[Devlin(2018)]%
        {devlin2018bert}
\bibfield{author}{\bibinfo{person}{Jacob Devlin}.} \bibinfo{year}{2018}\natexlab{}.
\newblock \showarticletitle{Bert: Pre-training of deep bidirectional transformers for language understanding}.
\newblock \bibinfo{journal}{\emph{arXiv preprint arXiv:1810.04805}} (\bibinfo{year}{2018}).
\newblock


\bibitem[Fei et~al\mbox{.}(2023)]%
        {fei2023lawbench}
\bibfield{author}{\bibinfo{person}{Zhiwei Fei}, \bibinfo{person}{Xiaoyu Shen}, \bibinfo{person}{Dawei Zhu}, \bibinfo{person}{Fengzhe Zhou}, \bibinfo{person}{Zhuo Han}, \bibinfo{person}{Songyang Zhang}, \bibinfo{person}{Kai Chen}, \bibinfo{person}{Zongwen Shen}, {and} \bibinfo{person}{Jidong Ge}.} \bibinfo{year}{2023}\natexlab{}.
\newblock \showarticletitle{Lawbench: Benchmarking legal knowledge of large language models}.
\newblock \bibinfo{journal}{\emph{arXiv preprint arXiv:2309.16289}} (\bibinfo{year}{2023}).
\newblock


\bibitem[Gao et~al\mbox{.}(2023)]%
        {gao2023retrieval}
\bibfield{author}{\bibinfo{person}{Yunfan Gao}, \bibinfo{person}{Yun Xiong}, \bibinfo{person}{Xinyu Gao}, \bibinfo{person}{Kangxiang Jia}, \bibinfo{person}{Jinliu Pan}, \bibinfo{person}{Yuxi Bi}, \bibinfo{person}{Yi Dai}, \bibinfo{person}{Jiawei Sun}, {and} \bibinfo{person}{Haofen Wang}.} \bibinfo{year}{2023}\natexlab{}.
\newblock \showarticletitle{Retrieval-augmented generation for large language models: A survey}.
\newblock \bibinfo{journal}{\emph{arXiv preprint arXiv:2312.10997}} (\bibinfo{year}{2023}).
\newblock


\bibitem[He et~al\mbox{.}({[n.\,d.]})]%
        {herevisiting}
\bibfield{author}{\bibinfo{person}{Junxian He}, \bibinfo{person}{Jiatao Gu}, \bibinfo{person}{Jiajun Shen}, {and} \bibinfo{person}{Marc'Aurelio Ranzato}.} \bibinfo{year}{[n.\,d.]}\natexlab{}.
\newblock \showarticletitle{Revisiting Self-Training for Neural Sequence Generation}. In \bibinfo{booktitle}{\emph{International Conference on Learning Representations}}.
\newblock


\bibitem[He et~al\mbox{.}(2023)]%
        {HanFei}
\bibfield{author}{\bibinfo{person}{Wanwei He}, \bibinfo{person}{Jiabao Wen}, \bibinfo{person}{Lei Zhang}, \bibinfo{person}{Hao Cheng}, \bibinfo{person}{Bowen Qin}, \bibinfo{person}{Yunshui Li}, \bibinfo{person}{Feng Jiang}, \bibinfo{person}{Junying Chen}, \bibinfo{person}{Benyou Wang}, {and} \bibinfo{person}{Min Yang}.} \bibinfo{year}{2023}\natexlab{}.
\newblock \bibinfo{title}{HanFei-1.0}.
\newblock \bibinfo{howpublished}{\url{https://github.com/siat-nlp/HanFei}}.
\newblock


\bibitem[Hu et~al\mbox{.}(2021)]%
        {hu2021lora}
\bibfield{author}{\bibinfo{person}{Edward~J Hu}, \bibinfo{person}{Yelong Shen}, \bibinfo{person}{Phillip Wallis}, \bibinfo{person}{Zeyuan Allen-Zhu}, \bibinfo{person}{Yuanzhi Li}, \bibinfo{person}{Shean Wang}, \bibinfo{person}{Lu Wang}, {and} \bibinfo{person}{Weizhu Chen}.} \bibinfo{year}{2021}\natexlab{}.
\newblock \showarticletitle{Lora: Low-rank adaptation of large language models}.
\newblock \bibinfo{journal}{\emph{arXiv preprint arXiv:2106.09685}} (\bibinfo{year}{2021}).
\newblock


\bibitem[Hurst et~al\mbox{.}(2024)]%
        {hurst2024gpt}
\bibfield{author}{\bibinfo{person}{Aaron Hurst}, \bibinfo{person}{Adam Lerer}, \bibinfo{person}{Adam~P Goucher}, \bibinfo{person}{Adam Perelman}, \bibinfo{person}{Aditya Ramesh}, \bibinfo{person}{Aidan Clark}, \bibinfo{person}{AJ Ostrow}, \bibinfo{person}{Akila Welihinda}, \bibinfo{person}{Alan Hayes}, \bibinfo{person}{Alec Radford}, {et~al\mbox{.}}} \bibinfo{year}{2024}\natexlab{}.
\newblock \showarticletitle{Gpt-4o system card}.
\newblock \bibinfo{journal}{\emph{arXiv preprint arXiv:2410.21276}} (\bibinfo{year}{2024}).
\newblock


\bibitem[Jiang and Yang(2023)]%
        {jiang2023legal}
\bibfield{author}{\bibinfo{person}{Cong Jiang} {and} \bibinfo{person}{Xiaolei Yang}.} \bibinfo{year}{2023}\natexlab{}.
\newblock \showarticletitle{Legal syllogism prompting: Teaching large language models for legal judgment prediction}. In \bibinfo{booktitle}{\emph{Proceedings of the Nineteenth International Conference on Artificial Intelligence and Law}}. \bibinfo{pages}{417--421}.
\newblock


\bibitem[Jiang et~al\mbox{.}(2023)]%
        {jiang2023longllmlingua}
\bibfield{author}{\bibinfo{person}{Huiqiang Jiang}, \bibinfo{person}{Qianhui Wu}, \bibinfo{person}{Xufang Luo}, \bibinfo{person}{Dongsheng Li}, \bibinfo{person}{Chin-Yew Lin}, \bibinfo{person}{Yuqing Yang}, {and} \bibinfo{person}{Lili Qiu}.} \bibinfo{year}{2023}\natexlab{}.
\newblock \showarticletitle{Longllmlingua: Accelerating and enhancing llms in long context scenarios via prompt compression}.
\newblock \bibinfo{journal}{\emph{arXiv preprint arXiv:2310.06839}} (\bibinfo{year}{2023}).
\newblock


\bibitem[Kojima et~al\mbox{.}(2022)]%
        {kojima2022large}
\bibfield{author}{\bibinfo{person}{Takeshi Kojima}, \bibinfo{person}{Shixiang~Shane Gu}, \bibinfo{person}{Machel Reid}, \bibinfo{person}{Yutaka Matsuo}, {and} \bibinfo{person}{Yusuke Iwasawa}.} \bibinfo{year}{2022}\natexlab{}.
\newblock \showarticletitle{Large language models are zero-shot reasoners}.
\newblock \bibinfo{journal}{\emph{Advances in neural information processing systems}}  \bibinfo{volume}{35} (\bibinfo{year}{2022}), \bibinfo{pages}{22199--22213}.
\newblock


\bibitem[Kullback and Leibler(1951)]%
        {kullback1951information}
\bibfield{author}{\bibinfo{person}{Solomon Kullback} {and} \bibinfo{person}{Richard~A Leibler}.} \bibinfo{year}{1951}\natexlab{}.
\newblock \showarticletitle{On information and sufficiency}.
\newblock \bibinfo{journal}{\emph{The annals of mathematical statistics}} \bibinfo{volume}{22}, \bibinfo{number}{1} (\bibinfo{year}{1951}), \bibinfo{pages}{79--86}.
\newblock


\bibitem[Lewis et~al\mbox{.}(2020)]%
        {lewis2020retrieval}
\bibfield{author}{\bibinfo{person}{Patrick Lewis}, \bibinfo{person}{Ethan Perez}, \bibinfo{person}{Aleksandra Piktus}, \bibinfo{person}{Fabio Petroni}, \bibinfo{person}{Vladimir Karpukhin}, \bibinfo{person}{Naman Goyal}, \bibinfo{person}{Heinrich K{\"u}ttler}, \bibinfo{person}{Mike Lewis}, \bibinfo{person}{Wen-tau Yih}, \bibinfo{person}{Tim Rockt{\"a}schel}, {et~al\mbox{.}}} \bibinfo{year}{2020}\natexlab{}.
\newblock \showarticletitle{Retrieval-augmented generation for knowledge-intensive nlp tasks}.
\newblock \bibinfo{journal}{\emph{Advances in Neural Information Processing Systems}}  \bibinfo{volume}{33} (\bibinfo{year}{2020}), \bibinfo{pages}{9459--9474}.
\newblock


\bibitem[Li et~al\mbox{.}(2024)]%
        {li2024lexevalcomprehensivechineselegal}
\bibfield{author}{\bibinfo{person}{Haitao Li}, \bibinfo{person}{You Chen}, \bibinfo{person}{Qingyao Ai}, \bibinfo{person}{Yueyue Wu}, \bibinfo{person}{Ruizhe Zhang}, {and} \bibinfo{person}{Yiqun Liu}.} \bibinfo{year}{2024}\natexlab{}.
\newblock \bibinfo{title}{LexEval: A Comprehensive Chinese Legal Benchmark for Evaluating Large Language Models}.
\newblock
\newblock
\showeprint[arxiv]{2409.20288}~[cs.CL]
\urldef\tempurl%
\url{https://arxiv.org/abs/2409.20288}
\showURL{%
\tempurl}


\bibitem[Li et~al\mbox{.}({[n.\,d.]})]%
        {lilexeval}
\bibfield{author}{\bibinfo{person}{Haitao Li}, \bibinfo{person}{You Chen}, \bibinfo{person}{Qingyao Ai}, \bibinfo{person}{WU Yueyue}, \bibinfo{person}{Ruizhe Zhang}, {and} \bibinfo{person}{LIU Yiqun}.} \bibinfo{year}{[n.\,d.]}\natexlab{}.
\newblock \showarticletitle{LexEval: A Comprehensive Chinese Legal Benchmark for Evaluating Large Language Models}. In \bibinfo{booktitle}{\emph{The Thirty-eight Conference on Neural Information Processing Systems Datasets and Benchmarks Track}}.
\newblock


\bibitem[Li et~al\mbox{.}(2023)]%
        {li2023compressing}
\bibfield{author}{\bibinfo{person}{Yucheng Li}, \bibinfo{person}{Bo Dong}, \bibinfo{person}{Frank Guerin}, {and} \bibinfo{person}{Chenghua Lin}.} \bibinfo{year}{2023}\natexlab{}.
\newblock \showarticletitle{Compressing Context to Enhance Inference Efficiency of Large Language Models}. In \bibinfo{booktitle}{\emph{Proceedings of the 2023 Conference on Empirical Methods in Natural Language Processing}}. \bibinfo{pages}{6342--6353}.
\newblock


\bibitem[Lin(2004)]%
        {lin2004rouge}
\bibfield{author}{\bibinfo{person}{Chin-Yew Lin}.} \bibinfo{year}{2004}\natexlab{}.
\newblock \showarticletitle{Rouge: A package for automatic evaluation of summaries}. In \bibinfo{booktitle}{\emph{Text summarization branches out}}. \bibinfo{pages}{74--81}.
\newblock


\bibitem[Liu et~al\mbox{.}(2023)]%
        {LAWGPT-zh}
\bibfield{author}{\bibinfo{person}{Hongcheng Liu}, \bibinfo{person}{Yusheng Liao}, \bibinfo{person}{Yutong Meng}, {and} \bibinfo{person}{Yuhao Wang}.} \bibinfo{year}{2023}\natexlab{}.
\newblock \bibinfo{title}{XieZhi: Chinese Law Large Language Model}.
\newblock \bibinfo{howpublished}{\url{https://github.com/LiuHC0428/LAW_GPT}}.
\newblock


\bibitem[Louis et~al\mbox{.}(2024)]%
        {louis2024interpretable}
\bibfield{author}{\bibinfo{person}{Antoine Louis}, \bibinfo{person}{Gijs van Dijck}, {and} \bibinfo{person}{Gerasimos Spanakis}.} \bibinfo{year}{2024}\natexlab{}.
\newblock \showarticletitle{Interpretable long-form legal question answering with retrieval-augmented large language models}. In \bibinfo{booktitle}{\emph{Proceedings of the AAAI Conference on Artificial Intelligence}}, Vol.~\bibinfo{volume}{38}. \bibinfo{pages}{22266--22275}.
\newblock


\bibitem[Luong et~al\mbox{.}(2024)]%
        {luong2024reft}
\bibfield{author}{\bibinfo{person}{Trung~Quoc Luong}, \bibinfo{person}{Xinbo Zhang}, \bibinfo{person}{Zhanming Jie}, \bibinfo{person}{Peng Sun}, \bibinfo{person}{Xiaoran Jin}, {and} \bibinfo{person}{Hang Li}.} \bibinfo{year}{2024}\natexlab{}.
\newblock \showarticletitle{Reft: Reasoning with reinforced fine-tuning}.
\newblock \bibinfo{journal}{\emph{arXiv preprint arXiv:2401.08967}} (\bibinfo{year}{2024}).
\newblock


\bibitem[Papineni et~al\mbox{.}(2002)]%
        {papineni2002bleu}
\bibfield{author}{\bibinfo{person}{Kishore Papineni}, \bibinfo{person}{Salim Roukos}, \bibinfo{person}{Todd Ward}, {and} \bibinfo{person}{Wei-Jing Zhu}.} \bibinfo{year}{2002}\natexlab{}.
\newblock \showarticletitle{Bleu: a method for automatic evaluation of machine translation}. In \bibinfo{booktitle}{\emph{Proceedings of the 40th annual meeting of the Association for Computational Linguistics}}. \bibinfo{pages}{311--318}.
\newblock


\bibitem[Reimers(2019)]%
        {reimers2019sentence}
\bibfield{author}{\bibinfo{person}{N Reimers}.} \bibinfo{year}{2019}\natexlab{}.
\newblock \showarticletitle{Sentence-BERT: Sentence Embeddings using Siamese BERT-Networks}.
\newblock \bibinfo{journal}{\emph{arXiv preprint arXiv:1908.10084}} (\bibinfo{year}{2019}).
\newblock


\bibitem[Robertson et~al\mbox{.}(1995)]%
        {robertson1995okapi}
\bibfield{author}{\bibinfo{person}{Stephen~E Robertson}, \bibinfo{person}{Steve Walker}, \bibinfo{person}{Susan Jones}, \bibinfo{person}{Micheline~M Hancock-Beaulieu}, \bibinfo{person}{Mike Gatford}, {et~al\mbox{.}}} \bibinfo{year}{1995}\natexlab{}.
\newblock \showarticletitle{Okapi at TREC-3}.
\newblock \bibinfo{journal}{\emph{Nist Special Publication Sp}}  \bibinfo{volume}{109} (\bibinfo{year}{1995}), \bibinfo{pages}{109}.
\newblock


\bibitem[Schulman et~al\mbox{.}(2017)]%
        {schulman2017proximal}
\bibfield{author}{\bibinfo{person}{John Schulman}, \bibinfo{person}{Filip Wolski}, \bibinfo{person}{Prafulla Dhariwal}, \bibinfo{person}{Alec Radford}, {and} \bibinfo{person}{Oleg Klimov}.} \bibinfo{year}{2017}\natexlab{}.
\newblock \showarticletitle{Proximal policy optimization algorithms}.
\newblock \bibinfo{journal}{\emph{arXiv preprint arXiv:1707.06347}} (\bibinfo{year}{2017}).
\newblock


\bibitem[Shannon(1948)]%
        {shannon1948mathematical}
\bibfield{author}{\bibinfo{person}{Claude~Elwood Shannon}.} \bibinfo{year}{1948}\natexlab{}.
\newblock \showarticletitle{A mathematical theory of communication}.
\newblock \bibinfo{journal}{\emph{The Bell system technical journal}} \bibinfo{volume}{27}, \bibinfo{number}{3} (\bibinfo{year}{1948}), \bibinfo{pages}{379--423}.
\newblock


\bibitem[Shao et~al\mbox{.}(2023)]%
        {shao2023enhancing}
\bibfield{author}{\bibinfo{person}{Zhihong Shao}, \bibinfo{person}{Yeyun Gong}, \bibinfo{person}{Yelong Shen}, \bibinfo{person}{Minlie Huang}, \bibinfo{person}{Nan Duan}, {and} \bibinfo{person}{Weizhu Chen}.} \bibinfo{year}{2023}\natexlab{}.
\newblock \showarticletitle{Enhancing Retrieval-Augmented Large Language Models with Iterative Retrieval-Generation Synergy}. In \bibinfo{booktitle}{\emph{Findings of the Association for Computational Linguistics: EMNLP 2023}}. \bibinfo{pages}{9248--9274}.
\newblock


\bibitem[Shum et~al\mbox{.}(2023)]%
        {shum2023automatic}
\bibfield{author}{\bibinfo{person}{Kashun Shum}, \bibinfo{person}{Shizhe Diao}, {and} \bibinfo{person}{Tong Zhang}.} \bibinfo{year}{2023}\natexlab{}.
\newblock \showarticletitle{Automatic Prompt Augmentation and Selection with Chain-of-Thought from Labeled Data}. In \bibinfo{booktitle}{\emph{Findings of the Association for Computational Linguistics: EMNLP 2023}}. \bibinfo{pages}{12113--12139}.
\newblock


\bibitem[Trautmann et~al\mbox{.}(2022)]%
        {trautmann2022legal}
\bibfield{author}{\bibinfo{person}{Dietrich Trautmann}, \bibinfo{person}{Alina Petrova}, {and} \bibinfo{person}{Frank Schilder}.} \bibinfo{year}{2022}\natexlab{}.
\newblock \showarticletitle{Legal prompt engineering for multilingual legal judgement prediction}.
\newblock \bibinfo{journal}{\emph{arXiv preprint arXiv:2212.02199}} (\bibinfo{year}{2022}).
\newblock


\bibitem[Wei et~al\mbox{.}(2022)]%
        {wei2022chain}
\bibfield{author}{\bibinfo{person}{Jason Wei}, \bibinfo{person}{Xuezhi Wang}, \bibinfo{person}{Dale Schuurmans}, \bibinfo{person}{Maarten Bosma}, \bibinfo{person}{Fei Xia}, \bibinfo{person}{Ed Chi}, \bibinfo{person}{Quoc~V Le}, \bibinfo{person}{Denny Zhou}, {et~al\mbox{.}}} \bibinfo{year}{2022}\natexlab{}.
\newblock \showarticletitle{Chain-of-thought prompting elicits reasoning in large language models}.
\newblock \bibinfo{journal}{\emph{Advances in neural information processing systems}}  \bibinfo{volume}{35} (\bibinfo{year}{2022}), \bibinfo{pages}{24824--24837}.
\newblock


\bibitem[Wu et~al\mbox{.}(2023)]%
        {sdu_fuzi_mingcha}
\bibfield{author}{\bibinfo{person}{Shiguang Wu}, \bibinfo{person}{Zhongkun Liu}, \bibinfo{person}{Zhen Zhang}, \bibinfo{person}{Zheng Chen}, \bibinfo{person}{Wentao Deng}, \bibinfo{person}{Wenhao Zhang}, \bibinfo{person}{Jiyuan Yang}, \bibinfo{person}{Zhitao Yao}, \bibinfo{person}{Yougang Lyu}, \bibinfo{person}{Xin Xin}, \bibinfo{person}{Shen Gao}, \bibinfo{person}{Pengjie Ren}, \bibinfo{person}{Zhaochun Ren}, {and} \bibinfo{person}{Zhumin Chen}.} \bibinfo{year}{2023}\natexlab{}.
\newblock \bibinfo{booktitle}{\emph{{fuzi.mingcha}}}.
\newblock


\bibitem[Wu et~al\mbox{.}({[n.\,d.]})]%
        {wisdomInterrogatory}
\bibfield{author}{\bibinfo{person}{Yiquan Wu}, \bibinfo{person}{Yuhang Liu}, \bibinfo{person}{Yifei Liu}, \bibinfo{person}{Ang Li}, \bibinfo{person}{Siying Zhou}, {and} \bibinfo{person}{Kun Kuang}.} \bibinfo{year}{[n.\,d.]}\natexlab{}.
\newblock \bibinfo{booktitle}{\emph{wisdomInterrogatory}}.
\newblock
\urldef\tempurl%
\url{https://github.com/zhihaiLLM/wisdomInterrogatory}
\showURL{%
\tempurl}
\newblock
\shownote{Available at GitHub}.


\bibitem[Xiao et~al\mbox{.}(2023)]%
        {bge_embedding}
\bibfield{author}{\bibinfo{person}{Shitao Xiao}, \bibinfo{person}{Zheng Liu}, \bibinfo{person}{Peitian Zhang}, {and} \bibinfo{person}{Niklas Muennighoff}.} \bibinfo{year}{2023}\natexlab{}.
\newblock \bibinfo{title}{C-Pack: Packaged Resources To Advance General Chinese Embedding}.
\newblock
\newblock
\showeprint[arxiv]{2309.07597}~[cs.CL]


\bibitem[Yang et~al\mbox{.}(2024)]%
        {qwen2}
\bibfield{author}{\bibinfo{person}{An Yang}, \bibinfo{person}{Baosong Yang}, \bibinfo{person}{Binyuan Hui}, \bibinfo{person}{Bo Zheng}, \bibinfo{person}{Bowen Yu}, \bibinfo{person}{Chang Zhou}, \bibinfo{person}{Chengpeng Li}, \bibinfo{person}{Chengyuan Li}, \bibinfo{person}{Dayiheng Liu}, \bibinfo{person}{Fei Huang}, \bibinfo{person}{Guanting Dong}, \bibinfo{person}{Haoran Wei}, \bibinfo{person}{Huan Lin}, \bibinfo{person}{Jialong Tang}, \bibinfo{person}{Jialin Wang}, \bibinfo{person}{Jian Yang}, \bibinfo{person}{Jianhong Tu}, \bibinfo{person}{Jianwei Zhang}, \bibinfo{person}{Jianxin Ma}, \bibinfo{person}{Jin Xu}, \bibinfo{person}{Jingren Zhou}, \bibinfo{person}{Jinze Bai}, \bibinfo{person}{Jinzheng He}, \bibinfo{person}{Junyang Lin}, \bibinfo{person}{Kai Dang}, \bibinfo{person}{Keming Lu}, \bibinfo{person}{Keqin Chen}, \bibinfo{person}{Kexin Yang}, \bibinfo{person}{Mei Li}, \bibinfo{person}{Mingfeng Xue}, \bibinfo{person}{Na Ni}, \bibinfo{person}{Pei Zhang}, \bibinfo{person}{Peng Wang}, \bibinfo{person}{Ru
  Peng}, \bibinfo{person}{Rui Men}, \bibinfo{person}{Ruize Gao}, \bibinfo{person}{Runji Lin}, \bibinfo{person}{Shijie Wang}, \bibinfo{person}{Shuai Bai}, \bibinfo{person}{Sinan Tan}, \bibinfo{person}{Tianhang Zhu}, \bibinfo{person}{Tianhao Li}, \bibinfo{person}{Tianyu Liu}, \bibinfo{person}{Wenbin Ge}, \bibinfo{person}{Xiaodong Deng}, \bibinfo{person}{Xiaohuan Zhou}, \bibinfo{person}{Xingzhang Ren}, \bibinfo{person}{Xinyu Zhang}, \bibinfo{person}{Xipin Wei}, \bibinfo{person}{Xuancheng Ren}, \bibinfo{person}{Yang Fan}, \bibinfo{person}{Yang Yao}, \bibinfo{person}{Yichang Zhang}, \bibinfo{person}{Yu Wan}, \bibinfo{person}{Yunfei Chu}, \bibinfo{person}{Yuqiong Liu}, \bibinfo{person}{Zeyu Cui}, \bibinfo{person}{Zhenru Zhang}, {and} \bibinfo{person}{Zhihao Fan}.} \bibinfo{year}{2024}\natexlab{}.
\newblock \showarticletitle{Qwen2 Technical Report}.
\newblock \bibinfo{journal}{\emph{arXiv preprint arXiv:2407.10671}} (\bibinfo{year}{2024}).
\newblock


\bibitem[Yu et~al\mbox{.}(2022)]%
        {yu2022legal}
\bibfield{author}{\bibinfo{person}{Fangyi Yu}, \bibinfo{person}{Lee Quartey}, {and} \bibinfo{person}{Frank Schilder}.} \bibinfo{year}{2022}\natexlab{}.
\newblock \showarticletitle{Legal prompting: Teaching a language model to think like a lawyer}.
\newblock \bibinfo{journal}{\emph{arXiv preprint arXiv:2212.01326}} (\bibinfo{year}{2022}).
\newblock


\bibitem[Yue et~al\mbox{.}(2023)]%
        {yue2023disclawllm}
\bibfield{author}{\bibinfo{person}{Shengbin Yue}, \bibinfo{person}{Wei Chen}, \bibinfo{person}{Siyuan Wang}, \bibinfo{person}{Bingxuan Li}, \bibinfo{person}{Chenchen Shen}, \bibinfo{person}{Shujun Liu}, \bibinfo{person}{Yuxuan Zhou}, \bibinfo{person}{Yao Xiao}, \bibinfo{person}{Song Yun}, \bibinfo{person}{Xuanjing Huang}, {and} \bibinfo{person}{Zhongyu Wei}.} \bibinfo{year}{2023}\natexlab{}.
\newblock \bibinfo{title}{DISC-LawLLM: Fine-tuning Large Language Models for Intelligent Legal Services}.
\newblock
\newblock
\showeprint[arxiv]{2309.11325}~[cs.CL]


\bibitem[Zhang et~al\mbox{.}(2024b)]%
        {zhang2024llama}
\bibfield{author}{\bibinfo{person}{Di Zhang}, \bibinfo{person}{Jianbo Wu}, \bibinfo{person}{Jingdi Lei}, \bibinfo{person}{Tong Che}, \bibinfo{person}{Jiatong Li}, \bibinfo{person}{Tong Xie}, \bibinfo{person}{Xiaoshui Huang}, \bibinfo{person}{Shufei Zhang}, \bibinfo{person}{Marco Pavone}, \bibinfo{person}{Yuqiang Li}, {et~al\mbox{.}}} \bibinfo{year}{2024}\natexlab{b}.
\newblock \showarticletitle{Llama-berry: Pairwise optimization for o1-like olympiad-level mathematical reasoning}.
\newblock \bibinfo{journal}{\emph{arXiv preprint arXiv:2410.02884}} (\bibinfo{year}{2024}).
\newblock


\bibitem[Zhang et~al\mbox{.}(2024c)]%
        {zhang2024citalaw}
\bibfield{author}{\bibinfo{person}{Kepu Zhang}, \bibinfo{person}{Weijie Yu}, \bibinfo{person}{Sunhao Dai}, {and} \bibinfo{person}{Jun Xu}.} \bibinfo{year}{2024}\natexlab{c}.
\newblock \showarticletitle{CitaLaw: Enhancing LLM with Citations in Legal Domain}.
\newblock \bibinfo{journal}{\emph{arXiv preprint arXiv:2412.14556}} (\bibinfo{year}{2024}).
\newblock


\bibitem[Zhang et~al\mbox{.}(2019)]%
        {zhang2019bertscore}
\bibfield{author}{\bibinfo{person}{Tianyi Zhang}, \bibinfo{person}{Varsha Kishore}, \bibinfo{person}{Felix Wu}, \bibinfo{person}{Kilian~Q Weinberger}, {and} \bibinfo{person}{Yoav Artzi}.} \bibinfo{year}{2019}\natexlab{}.
\newblock \showarticletitle{Bertscore: Evaluating text generation with bert}.
\newblock \bibinfo{journal}{\emph{arXiv preprint arXiv:1904.09675}} (\bibinfo{year}{2019}).
\newblock


\bibitem[Zhang et~al\mbox{.}(2024a)]%
        {zhang2024raft}
\bibfield{author}{\bibinfo{person}{Tianjun Zhang}, \bibinfo{person}{Shishir~G Patil}, \bibinfo{person}{Naman Jain}, \bibinfo{person}{Sheng Shen}, \bibinfo{person}{Matei Zaharia}, \bibinfo{person}{Ion Stoica}, {and} \bibinfo{person}{Joseph~E Gonzalez}.} \bibinfo{year}{2024}\natexlab{a}.
\newblock \showarticletitle{Raft: Adapting language model to domain specific rag}.
\newblock \bibinfo{journal}{\emph{arXiv preprint arXiv:2403.10131}} (\bibinfo{year}{2024}).
\newblock


\bibitem[Zhang et~al\mbox{.}({[n.\,d.]})]%
        {zhangautomatic}
\bibfield{author}{\bibinfo{person}{Zhuosheng Zhang}, \bibinfo{person}{Aston Zhang}, \bibinfo{person}{Mu Li}, {and} \bibinfo{person}{Alex Smola}.} \bibinfo{year}{[n.\,d.]}\natexlab{}.
\newblock \showarticletitle{Automatic Chain of Thought Prompting in Large Language Models}. In \bibinfo{booktitle}{\emph{The Eleventh International Conference on Learning Representations}}.
\newblock


\bibitem[Zheng et~al\mbox{.}(2023)]%
        {zheng2023secrets}
\bibfield{author}{\bibinfo{person}{Rui Zheng}, \bibinfo{person}{Shihan Dou}, \bibinfo{person}{Songyang Gao}, \bibinfo{person}{Yuan Hua}, \bibinfo{person}{Wei Shen}, \bibinfo{person}{Binghai Wang}, \bibinfo{person}{Yan Liu}, \bibinfo{person}{Senjie Jin}, \bibinfo{person}{Qin Liu}, \bibinfo{person}{Yuhao Zhou}, {et~al\mbox{.}}} \bibinfo{year}{2023}\natexlab{}.
\newblock \showarticletitle{Secrets of rlhf in large language models part i: Ppo}.
\newblock \bibinfo{journal}{\emph{arXiv preprint arXiv:2307.04964}} (\bibinfo{year}{2023}).
\newblock


\bibitem[Zhong et~al\mbox{.}(2024)]%
        {zhong2024evaluation}
\bibfield{author}{\bibinfo{person}{Tianyang Zhong}, \bibinfo{person}{Zhengliang Liu}, \bibinfo{person}{Yi Pan}, \bibinfo{person}{Yutong Zhang}, \bibinfo{person}{Yifan Zhou}, \bibinfo{person}{Shizhe Liang}, \bibinfo{person}{Zihao Wu}, \bibinfo{person}{Yanjun Lyu}, \bibinfo{person}{Peng Shu}, \bibinfo{person}{Xiaowei Yu}, {et~al\mbox{.}}} \bibinfo{year}{2024}\natexlab{}.
\newblock \showarticletitle{Evaluation of openai o1: Opportunities and challenges of agi}.
\newblock \bibinfo{journal}{\emph{arXiv preprint arXiv:2409.18486}} (\bibinfo{year}{2024}).
\newblock


\end{thebibliography}

\appendix

\end{document}